\documentclass{article}

\usepackage{microtype}
\usepackage{graphicx}
\usepackage[table]{xcolor}
\usepackage{tabularx}
\usepackage{booktabs}

\usepackage{bbold}
\usepackage{amsmath,amsthm,amssymb}
\usepackage{amsfonts}
\usepackage{nicefrac} 
\usepackage{cases}
\usepackage{bm}

\usepackage[utf8]{inputenc}
\usepackage[T1]{fontenc}
\usepackage{pifont}
\usepackage{xspace}
\usepackage{hyphenat} % handle hyphenations with \hyp{}
\usepackage{hyperref}       % hyperlinks
\usepackage{url}            % simple URL typesetting
\usepackage{nicefrac}       % compact symbols for 1/2, etc.
\usepackage{microtype}      % microtypography
\usepackage{thmtools,thm-restate}
\usepackage{graphicx}
\usepackage{subcaption} % figures next to each otehr 
\usepackage{cleveref}
\usepackage{color, soul} 	% for highlighting feedback during writing. 
\usepackage{booktabs}

\usepackage{thm-restate}

\newtheorem*{theorem*}{Theorem}

\newcommand{\R}{\mathbb{R}}

\newcommand{\tr}{\text{tr}}

\newcommand{\sqrtm}{\textsc{scipy.linalg.sqrtm}}
\usepackage{pgfplots}
\DeclareUnicodeCharacter{2212}{−}
\usepgfplotslibrary{groupplots,dateplot}
\usetikzlibrary{patterns,shapes.arrows}
\pgfplotsset{compat=newest}
\usepackage{hyperref}
\usepackage{cleveref}

\usepackage[export]{adjustbox}

\newcommand{\fid}{\text{FID}}
\newcommand{\gan}{\text{}}
\newcommand{\disc}{\text{D}}

\usepackage[accepted]{icml2021}

\icmltitlerunning{Backpropagating through FID}

\begin{document}

\twocolumn[
\icmltitle{Backpropagating through Fréchet Inception Distance}

\icmlsetsymbol{equal}{*}

\begin{icmlauthorlist}
\icmlauthor{Alexander Mathiasen}{equal,au}
\icmlauthor{Frederik Hvilsh\o j}{equal,au}
\end{icmlauthorlist}

\icmlaffiliation{au}{Department of Computer Science, Aarhus University, Denmark}

\icmlcorrespondingauthor{Alexander Mathiasen}{alexander.mathiasen@gmail.com}

\icmlkeywords{Machine Learning, ICML}

\vskip 0.3in
]

\printAffiliationsAndNotice{\icmlEqualContribution}

\begin{abstract}
The Fréchet Inception Distance (FID) has been used to \emph{evaluate} hundreds of~generative~models. 
We~introduce FastFID, which can efficiently \emph{train} generative models with FID as a loss~function. 
Using~FID as an additional loss for Generative Adversarial Networks improves their FID. 
\end{abstract}

\section{Introduction}
Generative modeling is a popular approach to unsupervised learning, with applications in, e.g., computer vision \cite{dcgan} and drug discovery \cite{moses}. 
A key difficulty for generative models is to evaluate their performance. 

In computer vision, generative models are evaluated with the Fréchet Inception Distance (FID) \cite{fid}. 
Inspired by the popularity of FID for \emph{evaluating} generative models, we explore whether generative models can be \emph{trained} using FID as a loss function. 

To optimize FID as a loss function, we backpropagate gradients through FID.  
While such backpropagation is possible with automatic differentiation, it is very \emph{slow}. 
In~some cases, it can increase training time by 10~days. 
We~surpass this issue with a new algorithm, FastFID, which allows \emph{fast} backpropagation through FID~(\Cref{sec:algo}).

\begin{figure}[h]
    \centering
    \includegraphics[width=0.48\textwidth]{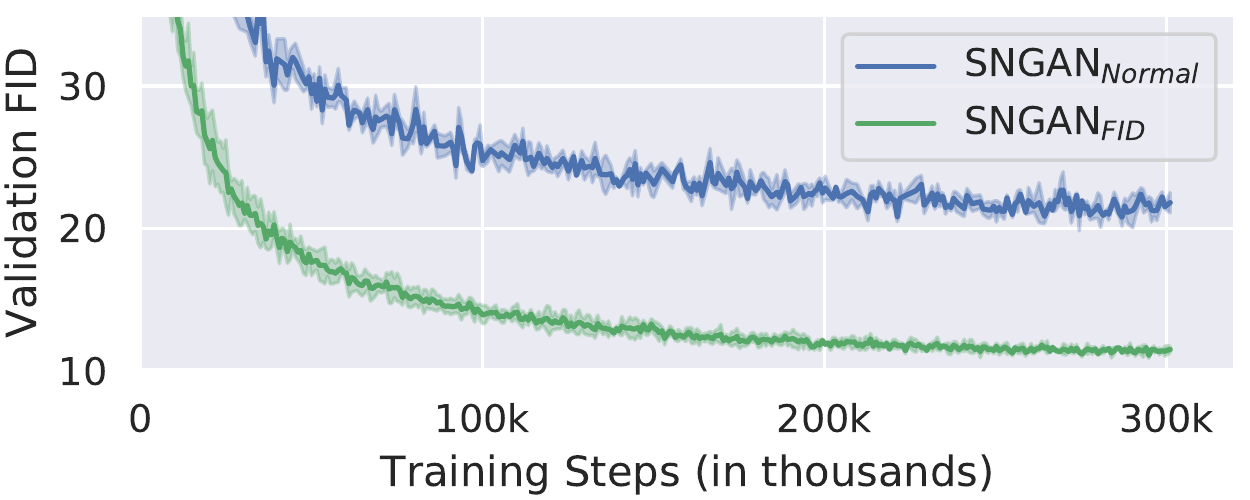}
    \vspace{-5mm}
    \caption{SNGAN trained normally (SNGAN$_\text{Normal}$) and SNGAN with FID as an additional loss (SNGAN$_\fid$). Lower FID is better. }
    \label{fig:frontpageloss}
\end{figure}

FastFID allows us to train Generative Adversarial Networks (GANs) \cite{gans} with FID as~a~loss. 
Training~GANs with FID as a loss improves validation FID. 
For~example, 
SNGAN \cite{sngan} gets 
FID 22 on CIFAR10 \cite{cifar}. 
If SNGAN uses FID as an additional loss, FID improves 
from~22~to~11, see~\Cref{fig:frontpageloss}. 
Interestingly,~SNGAN trained with FID beats several newer GANs, even though the newer GANs were improved with respect to both architecture and training. 
We consistently find similar improvements in FID across different GANs on different datasets (\Cref{sec:exp}). 

The improvements to FID raise an important question: Can optimization of FID as a loss ``improve'' generated~images?

\begin{figure}[h!]
    \centering
    \includegraphics[width=0.108\textwidth]{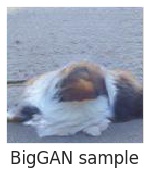}
    \hspace{2mm}
    \includegraphics[width=0.33\textwidth]{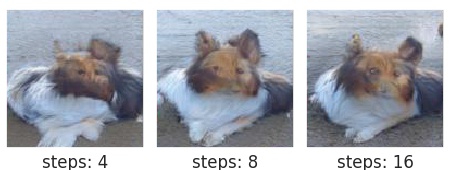}
    \caption{(Left) Fake image from pretrained BigGAN.~(Right) Training BigGAN with FID improves image by ``adding~a~head.'' }
    \label{fig:finetune}
\end{figure}

To answer this question, we take a pretrained BigGAN \cite{biggan} and train it to improve FID while inspecting the generated samples. 
We find that several samples improve with the addition of features like ears, eyes~or~tongues. 
For~example, one image of ``a dog without a head'' turns into an image of ``a dog with a head,'' see \Cref{fig:finetune}. 
Such examples demonstrate that training a generator to improve FID \emph{can} lead to better generated samples.
We present an analysis of FID as a loss function, which may explain the observed improvements
(\Cref{sec:finetune}). 

In conclusion, our work allows GANs to use FID as a loss, which improves FID and can improve generated images. 

The remainder of this paper is organized into sections~as follows.  
\Cref{sec:algo} introduces FastFID, a novel algorithm that supports fast backpropagation through FID. 
\Cref{sec:exp} demonstrates how three different GANs attain better validation FID when trained to explicitly minimize~FID. 
\Cref{sec:finetune}~explores whether optimizing FID as a loss can ``improve'' generated images. 

CODE: \verb^code.ipynb^ (one click to run in Google Colab).

\section{Fast Fréchet Inception Distance}\label{sec:algo}
%We first explain how FID \cite{fid} is computed. 
The FID between the model distribution $P_{model}$ and the real data distribution $P_{data}$ is computed as follows. 
Draw ``fake'' model samples $f_1,...,f_m\sim P_{model}$ and ``real'' data samples $r_1,...,r_n\sim P_{data}$.
Encode all samples $f_i$ and $r_i$ by computing activations $A(f_i)$ and $A(r_i)$ of the final layer of the pretrained Inception network~\cite{inception}. 
Compute the sample means $\mu_1,\mu_2$ and the sample covariance matrices $\Sigma_1, \Sigma_2$ of the activations $A(f_i), A(r_i)$. 
The FID is then the Wasserstein distance \cite{wasserstein} between the two multivariate normal distributions $N(\mu_1,\Sigma_1)$ and $N(\mu_2, \Sigma_2)$. 
\begin{equation}\label{equ:wasserstein}
    %W(N(\mu_1, \Sigma_1), N(\mu_2, \Sigma_2))^2= 
    ||\mu_1 - \mu_2||_2^2 + \tr(\Sigma_1) + \tr(\Sigma_2) - 2\cdot \tr( \sqrt{\Sigma_1\Sigma_2}).
\end{equation}
For evaluation during training, the original implementation use $10000$ samples by default \cite{fid}. 
On our workstation,\footnote{RTX 2080 Ti with Intel Xeon Silver 4214 CPU @ 2.20GHz, computing FID on CelebA \cite{celeba} images using precomputed Inception encodings of the real data. %FID implementation: \\ \href{https://github.com/hukkelas/pytorch-frechet-inception-distance}{https://github.com/hukkelas/pytorch-frechet-inception-distance}
(see appendix)
} this causes the FID evaluation to take approximately $20s$. 
Of the $20s$, it takes 
approximately $10s$ to compute Inception encodings and approximately $10s$ to compute $\tr(\sqrt{\Sigma_1\Sigma_2})$. 

The real data samples $r_1,...,r_n$ does not change \emph{during} training, so we only need to compute their Inception encodings once. 
The $10s$ spent computing the Inception encodings is only the time it takes to encode the fake model samples $f_1,...,f_m$. 
It is thus sufficient to reduce the number of fake samples $m$ to reduce the time spent computing Inception encodings.
For example, if we reduce $m$ from $10000$ to $128$ we reduce the time spent computing Inception encodings from $10s$ to $0.1s$. 

However, computing $\tr(\sqrt{\Sigma_1\Sigma_2})$ still takes $10s$. 
FastFID mitigates this issue by efficiently computing $\tr(\sqrt{\Sigma_1\Sigma_2})$ without explicitly computing $\sqrt{\Sigma_1\Sigma_2}$. 
\Cref{subsec:prevalgo} outlines how previous work computed $ \sqrt{\Sigma_1\Sigma_2}$ and \Cref{subsec:ouralgo} explains how FastFID computes $\tr(\sqrt{\Sigma_1\Sigma_2})$ efficiently. 

\subsection{Previous Algorithm } \label{subsec:prevalgo} % https://arxiv.org/pdf/1712.01034.pdf is O(d^^2 * convergence) 
The original FID implementation \cite{fid} computes $\tr(\sqrt{\Sigma_1\Sigma_2})$ by first constructing $\sqrt{\Sigma_1\Sigma_2}$. 
% The matrix square root is then computed using \verb^scipy.linalg.sqrtm^ \cite{scipy} which implements an extension of the algorithm from \cite{sqrtm} which is rich in matrix-matrix operations \cite{scipysqrtm}. 
The matrix square root is then computed using \sqrtm~ \cite{scipy} which implements an extension of the algorithm from \cite{sqrtm} which is rich in matrix-matrix operations \cite{scipysqrtm}. 
The algorithm starts by computing a Schur decomposition:
\begin{equation}\label{equ:schur}
    \Sigma_1\Sigma_2 = Q V Q^T\text{ with } Q^T=Q^{-1}\text{and triangular }V.
\end{equation}
The algorithm then computes a triangular matrix $U$ such that $U^2=V$ by using the triangular structure of $U$ and $V$. 

In particular, the triangular structure implies the following triangular equations $U_{ii}^2=V_{ii}$ and $U_{ii}U_{ij}+U_{ij}U_{jj}=V_{ij}-\sum_{k=i+1}^{j-1}U_{ik}U_{kj}$ \cite{scipysqrtm}. The equations can be solved wrt. $U$ one superdiagonal at a time. 
The resulting $U$ yields a matrix square root of the initial matrix. 
\begin{equation}
\label{equ:schur_sqrt}
    \sqrt{\Sigma_1\Sigma_2}=Q U Q^T.
\end{equation}
\paragraph{Time Complexity. }
Computing the Schur decomposition of $\Sigma_1\Sigma_2\in\mathbb{R}^{d\times d}$ takes $O(d^3)$ time. 
The resulting triangular equations can then be solved wrt. $U$ in $O(d^3)$ time. 
Computing~the matrix square root thus takes $O(d^3)$ time. 
FID~uses the Inception network which has $d=2048$. 
There exists other ``Fréchet-like'' distances, e.g. the Fréchet ChemNet Distance \cite{fcd} which uses the ChemNet network with $d=512$. 
%The size of $d$ depends on the network that encodes the data. 
%For example, FID uses the Inception network which has $d=2048$ while the Fréchet Chemnet Distance (FCD) \cite{fcd} uses the ChemNet network which has $d=512$. 
On our workstation, the different values of $d$ cause the square root computations to take approximately $10s$ for FID and $1s$ for FCD. % it is around 7s and .5s, I just think round gives less cognitive stain. 

\paragraph{Uniqueness. }
The matrix square root $\sqrt{M}$ is defined to be any matrix that satisfies $\sqrt{M}\sqrt{M}=M$. 
The square root of a matrix is in general not unique, i.e., some matrices have many square roots. 
The above algorithm does not necessarily find the same square root matrix if it is run several times, because 
the Schur decomposition is not unique. 
Furthermore, when computing $U_{ii}=\sqrt{T_{ii}}$ one has the freedom to choose both $\pm |\sqrt{T_{ii}}|$. The \sqrtm~ implementation chooses $U_{ii}=|\sqrt{T_{ii}}|$. 
Our implementation of the algorithm, presented in \Cref{subsec:ouralgo}, computes the trace $\tr(\sqrt{\Sigma_1\Sigma_2})$ such that it agrees with \sqrtm~ up to numerical errors.

\subsection{Our algorithm }\label{subsec:ouralgo}
This subsection presents an algorithm that computes $\tr(\sqrt{\Sigma_1\Sigma_2)}$ fast.
The high-level idea: construct a ``small'' matrix $M$ such that the eigenvalues $\lambda_i(M)$ satisfy $\sum_i|\sqrt{\lambda_i(M)}|=\tr(\sqrt{\Sigma_1\Sigma_2})$. 
Since $M$ is ``small,'' we can compute its eigenvalues faster than we can compute the matrix square root $\sqrt{\Sigma_1\Sigma_2}$. 

Let $X_1\in\R^{d\times m}$ have columns $A(f_1),...,A(f_m)$ and let $X_2\in\R^{d\times n}$ have columns $A(r_1),...,A(r_n)$. 
Let $\mathbb{1}_{k}$ be the $1\times k$ all ones vector. 
The sample covariance matrices $\Sigma_1$ and $\Sigma_2$ can then be computed as follows:
\begin{align}\label{equ:covmatrix1} 
    \Sigma_i= C_i C_i^T \quad\text{where}&\quad 
    C_1=\frac{1}{\sqrt{m-1}}(X_1-\mu_1 \mathbb{1}_{m}) \\ 
    \label{equ:covmatrix2} 
    \text{and}&\quad 
    C_2=\frac{1}{\sqrt{n-1}}(X_2-\mu_2 \mathbb{1}_{n}).
\end{align}
This allows us to write $\Sigma_1\Sigma_2=C_1C_1^T C_2C_2^T$. 
Recall that the eigenvalues of $AB$ are equal to the eigenvalues of $BA$ if both $AB$ and $BA$ are square matrices \cite{eigfact}. \clearpage 

The eigenvalues of the $d\times d$ matrix $C_1C_1^TC_2C_2^T$ are thus the same as the eigenvalues of the $m\times m$ matrix $M=C_1^TC_2C_2^TC_1$.
The matrix $M$ is small in the sense that we will use $m\ll d$, for example, for FID we often use $m=128$ fake samples while $d=2048$. 

We now show that $\sum_i|\sqrt{\lambda_i(M)}|=\tr(\sqrt{\Sigma_1\Sigma_2})$ if $\sqrt{\Sigma_1\Sigma_2}$ is computed by \sqrtm. 
Since $\tr(\sqrt{\Sigma_1\Sigma_2})=\sum_i\lambda_i(\sqrt{\Sigma_1\Sigma_2})$ it is sufficient to show that the eigenvalues of $\sqrt{\Sigma_1\Sigma_2}$ are equal to the positive square root of the eigenvalues of $\Sigma_1\Sigma_2$. 
In other words:
it is sufficient to show that $\lambda_i(\sqrt{\Sigma_1\Sigma_2})=|\sqrt{\lambda_i(\Sigma_1\Sigma_2)}|$.

Recall that $\sqrt{\Sigma_1\Sigma_2}=QUQ^T$ where $\Sigma_1\Sigma_2=QVQ^T$, $U^2=V$ and $U_{ii}^2=V_{ii}$. 
Since both $U$ and $V$ are triangular they have their eigenvalues on the diagonal, which means that $\lambda_i(\sqrt{\Sigma_1\Sigma_2})=U_{ii}=\sqrt{V_{ii}}=\sqrt{\lambda_i(\Sigma_1\Sigma_2)}$. 
Recall that \sqrtm~ chooses $U_{ii}=|\sqrt{V_{ii}}|$ and we get $\lambda_i(\sqrt{\Sigma_1\Sigma_2})=|\sqrt{\lambda_i(\Sigma_1\Sigma_2)}|$ as wanted. 
Note that even though the Schur decomposition $\Sigma_1\Sigma_2=QUQ^T$ is not unique, $U$ will always have the eigenvalues of $\Sigma_1\Sigma_2$ on its diagonal and thus preserve the trace. 
Putting everything together, we realize the desired result: 
$$
\tr(\sqrt{\Sigma_1\Sigma_2})=\sum_{i=1}^{m-1} |\sqrt{\lambda_i(C_1^TC_2C_2^TC_1)}|.$$
For completeness, we provide pseudo-code in \Cref{algo:pseudocode}.
The algorithm can be modified to compute FCD by simply changing the Inception network to the ChemNet network.

\begin{algorithm}[h!]
   \caption{Fast Fréchet Inception Distance}
   \label{algo:pseudocode}
\begin{algorithmic}[1]
   \STATE {\bfseries Input:} $f_1,...,f_m\sim P_{model}$, the Inception network $\text{Net}(x)$ and precomputed $\mu_2,C_2$. 
   \STATE 
   \STATE // Compute network activations 
   \STATE Compute $A(f_i)= \text{Net}(f_i)$ and let $X_1$ be a matrix with columns $A(f_1),...,A(f_m)$. %and $A(r_i) = \text{Net}(r_i)$
   %\COMMENT{\hfill $\triangleright \; O(dm^2)$} 
   \STATE 
   \STATE // Compute mean. 
   \STATE $\mu_1 = \frac{1}{m}\sum_{i=1}^m(X_1)_i $. 
   \STATE 
   \STATE // Compute trace of square root matrix. 
   \STATE $C_1=\frac{1}{\sqrt{m-1}}(X_1-\mu_1 \mathbb{1}_m)$  as in \Cref{equ:covmatrix1}. 
   \STATE $S = \text{library.eigenvalues}( (C_1^TC_2) (C_2^TC_1) )$
   \STATE $\tr(\sqrt{\Sigma_1\Sigma_2})= \sum_{i=1}^{m-1}|\sqrt{S_i}|$
   \STATE 
   \STATE // Compute trace of both covariance matrices. 
   \STATE $\tr(\Sigma_1)=\sum_{i=1}^m \text{row}_i(C_1)^T\text{row}_i(C_1)$ 
   \STATE $\tr(\Sigma_2)=\sum_{i=1}^n \text{row}_i(C_2)^T\text{row}_i(C_2)$
   \STATE 
   \STATE \textbf{return} $||\mu_1-\mu_2||_2^2 + \tr{(\Sigma_1)} + \tr{(\Sigma_2)} - 2 \cdot \tr(\sqrt{\Sigma_1\Sigma_2})$
\end{algorithmic}
\end{algorithm}

\paragraph{Time Complexity.}
Computing $M=(C_1^TC_2)(C_2^TC_1)$ takes $O(mdn+m^2n)$ time. 
The eigenvalues of $M$ can be computed in $O(m^3)$ time, giving a total time complexity of $O(mdn+m^2n+m^3)$. 
If we use a large number of real samples $n\gg d$, we can precompute $\Sigma_2=C_2C_2^T$ and compute $M=C_1^T\Sigma_2C_1$ in $O(d^2 m)$ time, giving a total time complexity of $O(d^2m+m^3)$. 

\paragraph{Computing Gradients. }
If the network $\text{Net}(x)$ used in \Cref{algo:pseudocode} supports  gradients with respect to its input, as the Inception network does, it is possible to compute gradients with respect to the input samples $f_i$. 
If $f_i$ were constructed by a generative model, one can also compute gradients wrt. the parameters of the generative model. 
If \Cref{algo:pseudocode} is implemented with automatic differentiation, e.g., through PyTorch \cite{pytorch} or TensorFlow \cite{tf}, these gradients are computed automatically. 
%For example, we implemented \Cref{algo:pseudocode} in PyTorch and used autograd to compute gradients when constructing adversarial examples in \Cref{sec:adv}. 

\paragraph{Potential Further Speed-ups. } 
\Cref{algo:pseudocode} computes $\tr(\sqrt{\Sigma_1\Sigma_2})$ fast. 
As a result, computing $\tr(\sqrt{\Sigma_1\Sigma_2})$ only takes a few percent of the total time spent by \Cref{algo:pseudocode}. 
The majority of the time spent by \Cref{algo:pseudocode} is spent computing Inception encodings. 
One could compute the encodings faster by compressing \cite{compress} the Inception network, at the cost of introducing some small~error.

\begin{figure*}[h]
    \centering
    \includegraphics[width=0.98\textwidth]{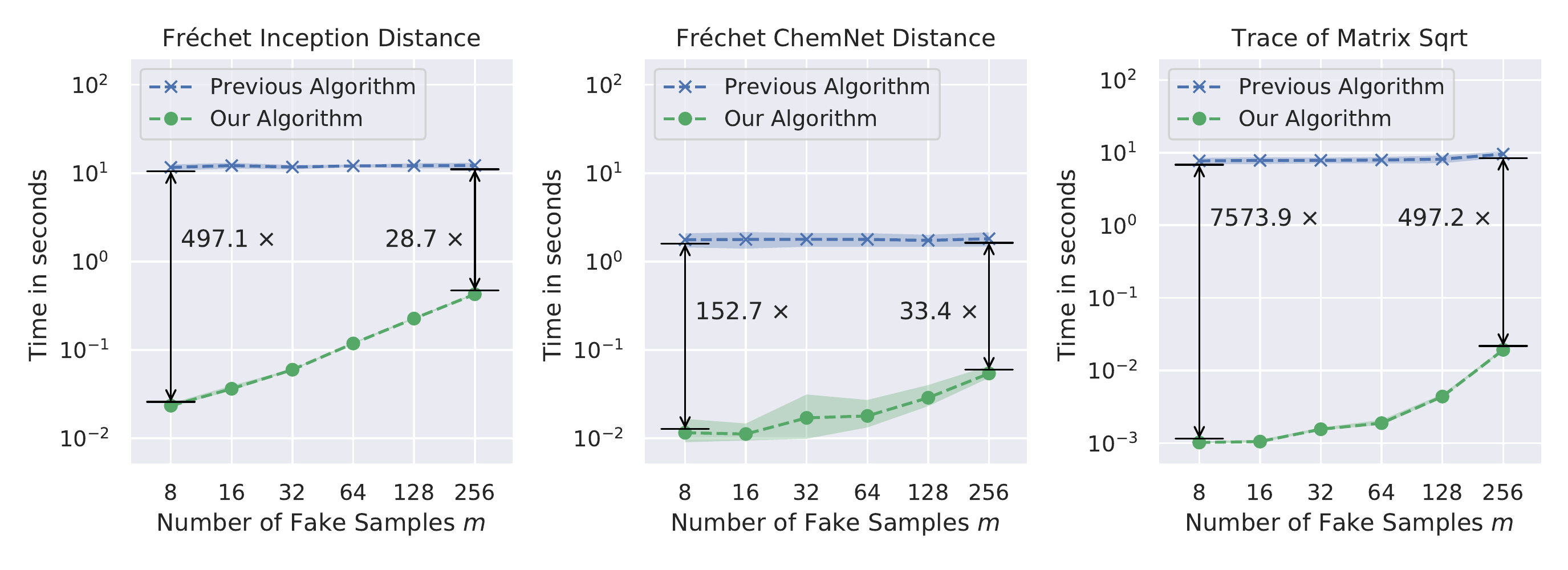}
    \caption{Comparison of \sqrtm~ and \Cref{algo:pseudocode} for different batch sizes. Left: Time to compute Fréchet Inception Distance. Middle: Time to compute Fréchet ChemNet Distance. Right: Time to compute the trace of matrix square root. %\textbf{todo: add time of svd of 2048 matrix, this is lowerbound of time for old algorithm if written for gpu; takes roughly 1+- 2 sec} 
    }
    \label{fig:speed}
\end{figure*}

\subsection{Experimental Speed-Up}\label{subsec:pracspeedup}
%In \Cref{sec:algo}, we presented a fast algorithm to compute $\tr(\sqrt{\Sigma_1\Sigma_2})$, which can be used to speed up the computation of both FID and FCD. 
In this subsection, we compare the running time of \Cref{algo:pseudocode} implemented in PyTorch \cite{pytorch} against open-source
implementations of FID, FCD and $\sqrt{\Sigma_1\Sigma_2}$. 
\footnote{\href{https://github.com/insilicomedicine/fcd_torch} {https://github.com/insilicomedicine/fcd\_torch}
\\\href{https://github.com/hukkelas/pytorch-frechet-inception-distance} {https://github.com/hukkelas/pytorch-frechet-inception-distance}
\\\href{https://github.com/scipy/scipy/blob/v1.6.0/scipy/linalg/_matfuncs_sqrtm.py\#L114-L191}{https://github.com/scipy/scipy/blob/v1.6.0/scipy/linalg/\_matfuncs\_sqrtm.py}
} 
The algorithms are compared as the number of different fake samples varies $m=8,16,\dots,256$. 

We used images from CelebA \cite{celeba} to time FID. 
We used molecules from MOSES \cite{moses} to time FCD. 
For both FID and FCD, we precomputed $\Sigma_2$ with $10000$ real~samples. 

To time $\tr(\sqrt{C_1^T\Sigma_2C_1})$, we used $\Sigma_2\in \R^{d\times d}$ with $N(0,1)$ entries. 
We computed $C_1$ by using \Cref{equ:covmatrix1} where $X_1\in \R^{d\times m}$ had $N(0,1)$ entries. 
To make the timing of $\tr(\sqrt{C_1^T\Sigma_2C_1})$ comparable with FID, we chose $d=2048$ to match the Inception network. 

For each $m$, we ran one warmup round and then repeated the experiment $100$ times. 
\Cref{fig:speed} plots average running time with standard deviation as error bars.\footnote{For our algorithm, we plot $[\mu-\sigma/2, \mu+\sigma]$ instead of $\mu\pm \sigma$ to avoid clutter caused by the logarithmic scaling. This means that the real running time of our algorithm is sometimes a little bit better than visualized. } 

The previous algorithms take roughly the same amount of time for all~$m$. 
This is expected since most of the time is spent computing $\sqrt{\Sigma_1\Sigma_2}$, which takes the same amount of time for all $m$.  
Our algorithm takes less time as $m$ decreases. 
This is expected since our algorithm takes $O(m^3+d^2m)$ time for precomputed $\Sigma_2$. 

For all $m$, our algorithm is at least $25$ times faster than the previous algorithms, and in some cases up to hundreds or even thousands of times faster. 
Notably, both algorithms computed the same thing, our algorithm just computed the same thing faster. 
To concretize the reduction in training time, we exemplify the reduction below. 

\paragraph{Example.} \Cref{sec:exp} trains GANs to minimize FID with batch size $m=128$. 
Our algorithm then reduces computation time from $12.15\pm 0.80$ seconds to $0.23\pm 0.01$ seconds (mean $\pm$ standard deviation). 
When training for $10^5$ steps (as done in \Cref{sec:exp}) this reduces time by 13 days. 
When training for $10^6$ steps (as done in \cite{sagan}) this reduces time by 130 days. % fix the we. 

%We emphasize that both algorithms computed the same thing, our algorithm just computed the same thing faster. 

% make this into a subsection 
\subsection{Numerical Error } \label{subsec:numerror}
To investigate the numerical error of \Cref{algo:pseudocode} and \sqrtm{}, we ran an experiment where the square root can easily be computed. 
If we choose $C_1=C_2$ then $C_1C_1^TC_2C_2^T=(C_1C_1^T)^2$ is positive semi-definite and has a unique positive semi-definite square root $\sqrt{(C_1C_1^T)^2}=C_1C_1^T$. 
Furthermore, it is exactly this positive semi-definite square root \sqrtm{} computes since the implementation chooses the positive eigenvalues $U_{ii}=|\sqrt{V_{ii}}|$ (see \Cref{subsec:prevalgo} for details). 
We~can~then investigate the numerical errors by comparing the ground truth $\tr(C_1C_1^T)$ with the result from both \Cref{algo:pseudocode} and \sqrtm{}. 

%\paragraph{TODO (only if time, if not, or need more space, just remove it, maybe to appendix)}: Both algo had low error at float 64-bit, so we tried float-32 and float16;; this allowed memory savings. %To amplify the numerical errors we used 32-bit floating points instead of 64-bit floating points. 

The experiment was repeated for $m=32,64,128,256$ number of fake samples, where $C_1$ was computed as done in \Cref{subsec:pracspeedup}. 
We report the ground truth $\tr(C_1C_1^T)$ and the absolute numerical error caused by \Cref{algo:pseudocode} and \sqrtm:
\begin{equation}
    \left|\tr(C_1C_1^T)-\tr(\textsc{scipy.sqrtm}(C_1C_1^TC_1C_1^T))\right|.
\end{equation}
See results in \Cref{fig:numerror}. 

\begin{table}[h]
    \centering
    \begin{tabular}{cccc}
    \toprule
         $m$ & Answer & Error \textsc{scipy} & Error \Cref{algo:pseudocode} \\
    \midrule
 8           & 14283& 175& 0.0000      \\
 %8           & 14283.8174  & 175.1039   & 0.0000      \\
    %\hline
 16          & 30678& 228& 0.0020      \\
 %16          & 30678.9355  & 228.1520   & 0.0020      \\
    %\hline
 32          & 62955& 300& 0.0000      \\
 %32          & 62955.6250  & 300.0112   & 0.0000      \\
    %\hline
 64          & 128947& 408& 0.0078     \\
 %64          & 128947.5625 & 408.3399   & 0.0078     \\
    %\hline
 128         & 259586& 565& 0.0156      \\
 %128         & 259586.1250 & 565.4011   & 0.0156      \\
    %\hline
 256         & 523360& 785& 0.0312     \\
% 256         & 523360.3438 & 785.0262   & 0.0312     \\
    \bottomrule
    \end{tabular}
    \caption{Numerical error of \textsc{scipy.sqrtm} and \Cref{algo:pseudocode} for different number of fake samples $m$. 
    We also add correct answer so the percentage wise numerical error can be inferred. 
    }
    \label{fig:numerror}
\end{table}

The numerical error of \Cref{algo:pseudocode} is at least $1000$ times smaller than that of \sqrtm. 
We suspect that \Cref{algo:pseudocode} has smaller numerical errors because it computes eigenvalues of a ``small'' $m\times m$ matrix instead~of computing a Schur decomposition of the ``full'' $d\times d$ matrix. 

The above experiment used 32 bit precision. 
By default, \sqrtm~ uses 64 bit precision and exhibit negligible numerical errors. 
Neural networks are usually trained with 32 bit precision and sometimes even 16 bit precision. 
Additional numerical stability is thus desirable as it allows us to reduce numerical precision. 

% TODO: add if asked for comparison. 
%\subsection{Approximate Iterative Methods }
%The matrix square root can be approximated as done by, e.g., \cite{approxsqrtm}. \footnote{\href{https://github.com/msubhransu/matrix-sqrt}{https://github.com/msubhransu/matrix-sqrt}}
%Such methods are much faster than \textsc{scipy.linalg.sqrtm} but introduce approximation errors. 
%For $d=2048$ it reduces time from roughly $10$s to $0.050$s. 
%Nonetheless, for batch-size $m=128$ as used in \Cref{sec:exp}, our algorithm takes $0.004$s and thus remains $\approx 12.5\times$ faster without approximation errors. 

\section{Training GANs with FID as a Loss}
\label{sec:exp} 
%This section demonstrates how FID training improves validation FID of GANs.  
This section demonstrates that GANs trained with FID as an additional  loss get better validation FID. %how FID training improves validation FID of GANs.  
Experimental~setup: 
we find an open-source implementation of a popular GAN, and train it while monitoring validation FID. 
We then train an identical GAN, but with FID as an additional~loss. 
The~experiment is repeated 3 times and we report $\mu\pm\sigma$ where $\mu$ is the mean validation FID and $\sigma$ is the standard deviation. 
For each repetition, we show 8 fake samples (see appendix in the supplementary material for enlarged images). 

To increase the robustness of our experiment, we performed the experiment with three different GANs on three different datasets. %
SNGAN \cite{sngan} on CIFAR10 \cite{cifar}, DCGAN \cite{dcgan} on CelebA \cite{celeba} and SAGAN \cite{sagan} on ImageNet \cite{imagenet}. 
To distinguish between GANs with and without FID loss we write, e.g., SNGAN$_\fid$ and SNGAN$_\gan$, respectively. 

\begin{figure}[h!]
    \centering
    \includegraphics[width=0.46\textwidth]{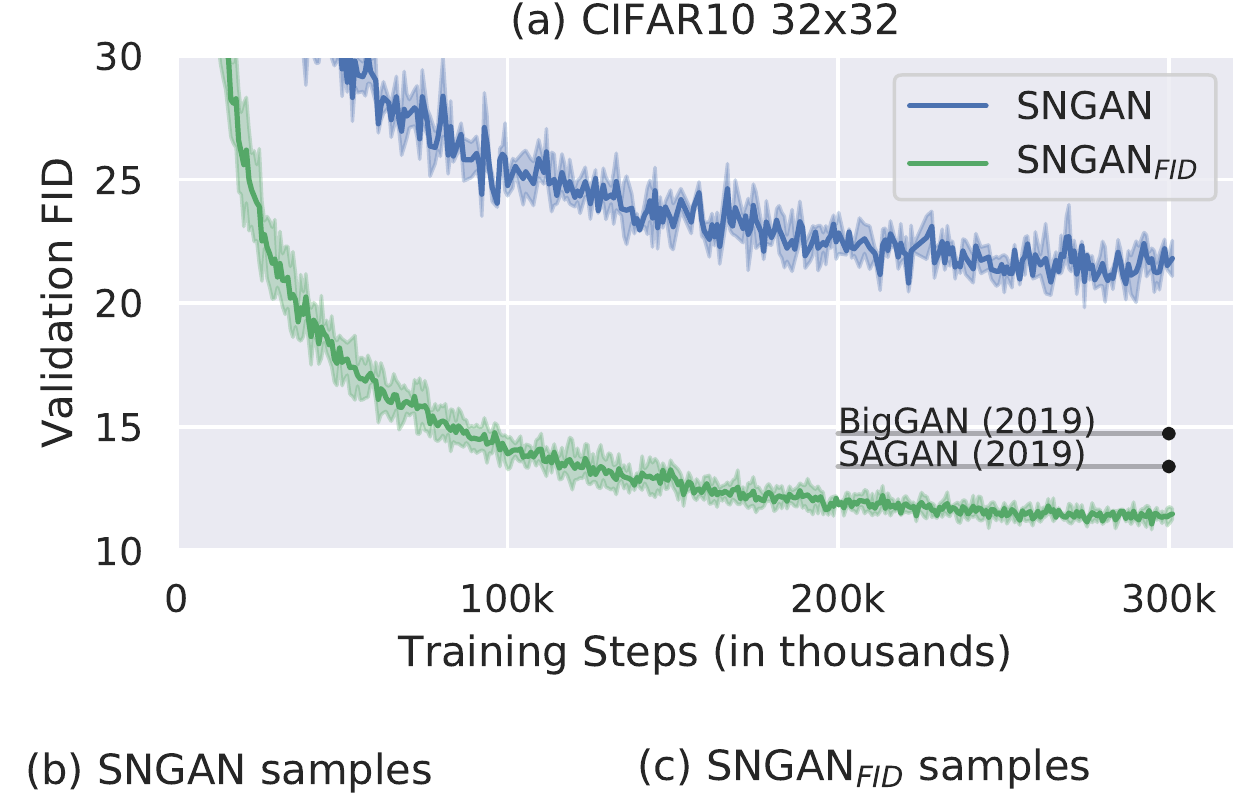}
    \adjincludegraphics[trim={{0.0\width} 0 0 0}, clip, width=0.23\textwidth]{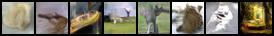}
    \adjincludegraphics[trim={{0.0\width} 0 0 0}, clip,width=0.23\textwidth]{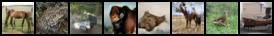}
    \adjincludegraphics[trim={{0.0\width} 0 0 0}, clip,width=0.23\textwidth]{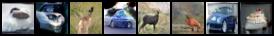}
    \adjincludegraphics[trim={{0.0\width} 0 0 0}, clip,width=0.23\textwidth]{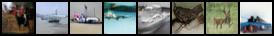}
    \adjincludegraphics[trim={{0.0\width} 0 0 0}, clip,width=0.23\textwidth]{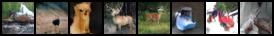}
    \adjincludegraphics[trim={{0.0\width} 0 0 0}, clip,width=0.23\textwidth]{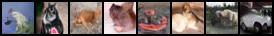}
    %\vspace*{-2mm}
    \caption{SNGAN trained on CIFAR10. (a) validation FID (b) fake samples from SNGAN$_\gan$ (c) fake samples from SNGAN$_\fid$.}
    \label{fig:sngan_train}
    %\vspace*{3.5mm}
%\end{figure}
%\begin{figure}[h]
    %\centering
    \includegraphics[width=0.48\textwidth]{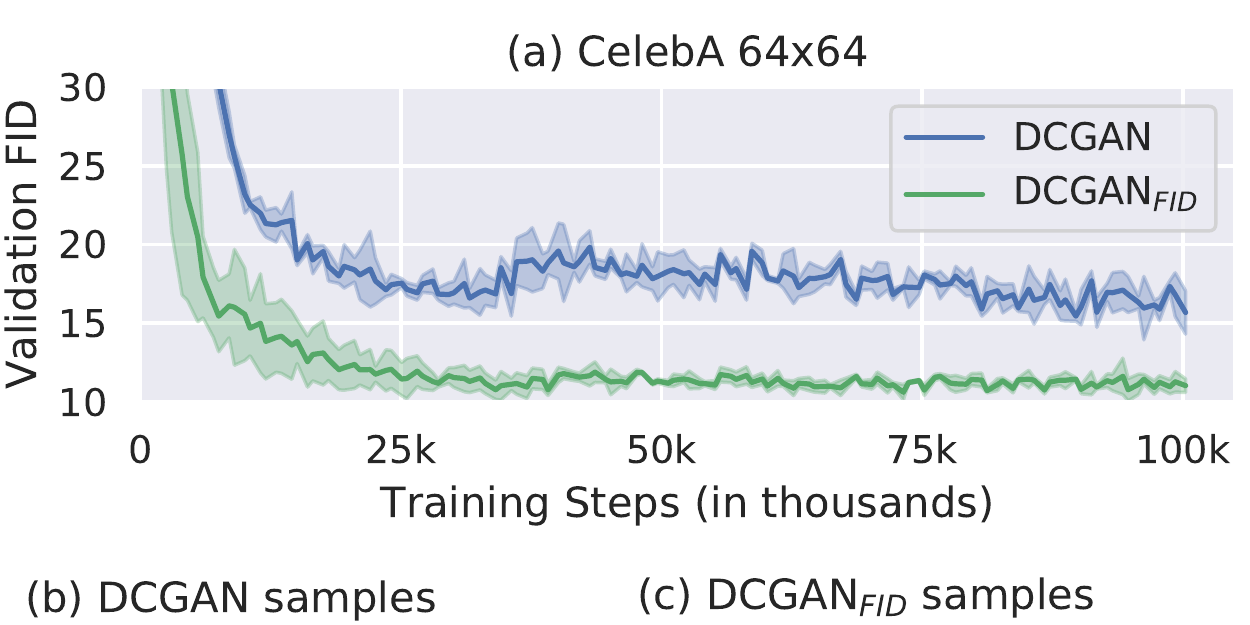}
    \adjincludegraphics[trim={{0.0\width} 0 0 0}, clip,width=0.23\textwidth]{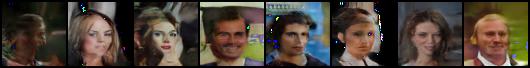}
    \adjincludegraphics[trim={{0.0\width} 0 0 0}, clip,width=0.23\textwidth]{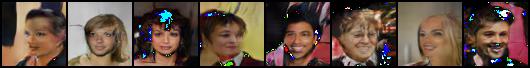}
    \adjincludegraphics[trim={{0.0\width} 0 0 0}, clip,width=0.23\textwidth]{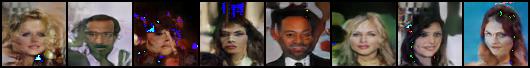}
    \adjincludegraphics[trim={{0.0\width} 0 0 0}, clip,width=0.23\textwidth]{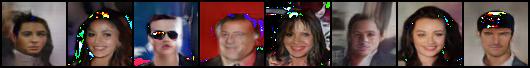}
    \adjincludegraphics[trim={{0.0\width} 0 0 0}, clip,width=0.23\textwidth]{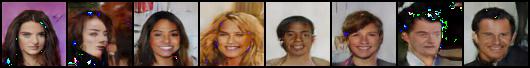}
    \adjincludegraphics[trim={{0.0\width} 0 0 0}, clip,width=0.23\textwidth]{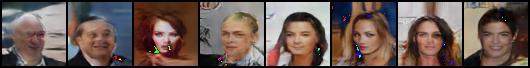}
    %\vspace*{-2mm}
   \caption{DCGAN trained on CelebA. (a) validation FID (b) fake samples from DCGAN$_\gan$ (c) fake samples from DCGAN$_\fid$.  }
    \label{fig:dcgan_train}
%\end{figure}
%\begin{figure}[h]
    %\centering
    %\vspace*{3.5mm}
    \includegraphics[width=0.48\textwidth]{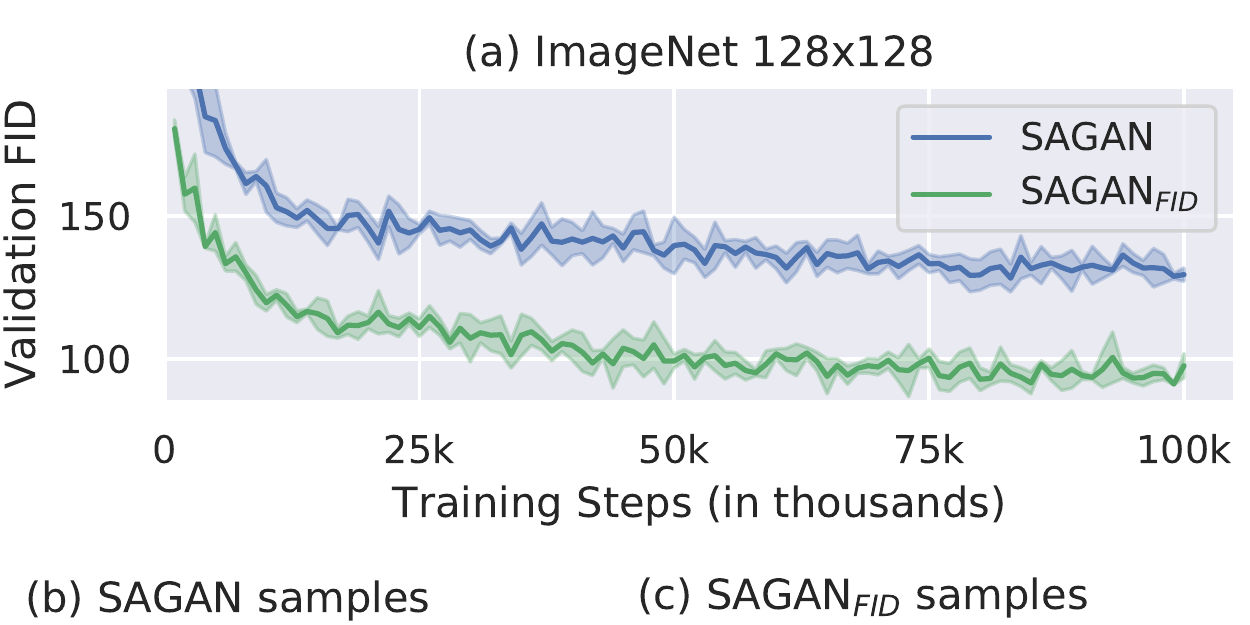}
    \adjincludegraphics[trim={{0.0\width} 0 0 0}, clip, width=0.23\textwidth]{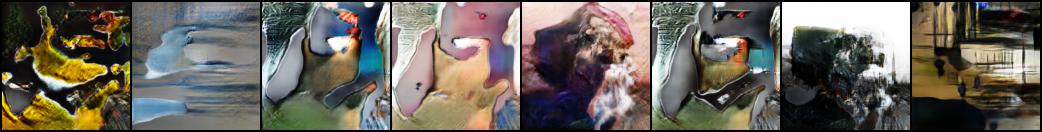}
    \adjincludegraphics[trim={{0.0\width} 0 0 0}, clip, width=0.23\textwidth]{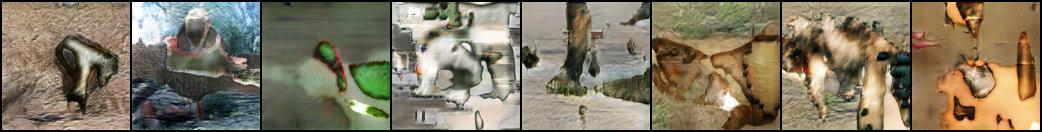}
    \adjincludegraphics[trim={{0.0\width} 0 0 2}, clip, width=0.23\textwidth]{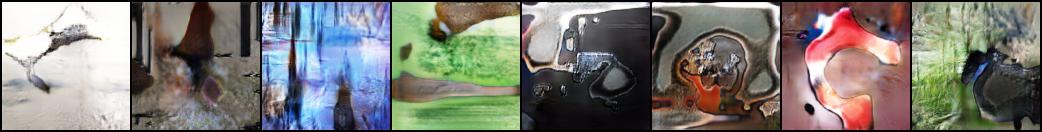}
    \adjincludegraphics[trim={{0.0\width} 0 0 0}, clip, width=0.23\textwidth]{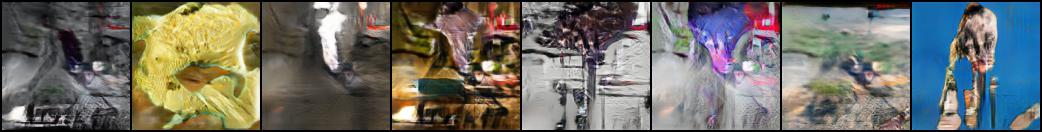}
    \adjincludegraphics[trim={{0.0\width} 0 0 0}, clip, width=0.23\textwidth]{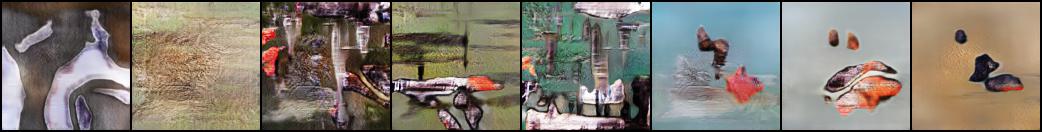}
    \adjincludegraphics[trim={{0.0\width} 0 0 0}, clip, width=0.23\textwidth]{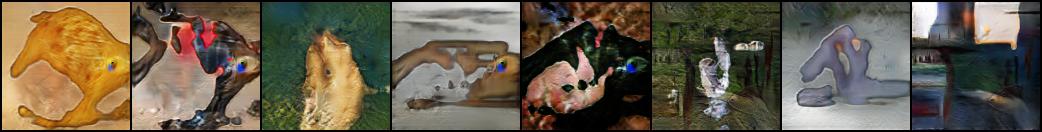}
    \caption{SAGAN trained on ImageNet. (a) validation FID (b) fake samples from SAGAN$_\gan$ (c) fake samples from SAGAN$_\fid$. }
    \label{fig:sagan_train}
\end{figure}

\textbf{SNGAN on CIFAR10.}
We train SNGAN\footnote{  \href{https://github.com/GongXinyuu/sngan.pytorch}{https://github.com/GongXinyuu/sngan.pytorch} } on CIFAR10. 
After 301070 training steps validation FID improved from $21.81\pm 0.73$ to $11.48\pm 0.22$. See \Cref{fig:sngan_train} for validation FID during training. 
Below the plot we include fake images from SAGAN$_\gan$ (left) and fake images from SAGAN$_\fid$ (right), each row corresponds to one repetition of the experiment. 

To add context, BigGAN and SAGAN got FID $14.73$ and $13.4$, respectively \cite{saganfid}.
Both SAGAN and BigGAN were published after SNGAN, and introduced improvements to both architecture and training. 
From this perspective, it is interesting that SNGAN can beat both SAGAN and BigGAN by simply adding FID to the loss function. 
State-of-the-art is 7.01 \cite{nvidia}.  

\textbf{DCGAN on CelebA.} We train DCGAN\footnote{  \href{https://github.com/Natsu6767/DCGAN-PyTorch}{https://github.com/Natsu6767/DCGAN-PyTorch}} on CelebA at 64x64 resolution. 
After 100264 training steps validation FID improved from $15.67\pm 1.38$ to $11.03\pm 0.43$. 
\Cref{fig:dcgan_train} contains validation FID and samples like \Cref{fig:sngan_train}. %for validation FID during training and samples . Below the plot we include fake images from the GAN (left) and fake images from the GAN augmented with FID (right), each row correspond to one repetition of the experiment.
%We found no signs of generator cheating. 

\textbf{SAGAN on ImageNet.} We train SAGAN\footnote{
 \href{https://github.com/rosinality/sagan-pytorch}{https://github.com/rosinality/sagan-pytorch}
} on ImageNet at 128x128 resolution. 
After 100001 training steps validation FID improved from $129.48\pm2.38$ to $97.64\pm 4.33$. 
\Cref{fig:sagan_train} contains validation FID and samples like \Cref{fig:sngan_train}. %previous figure. %for validation FID during training. Below the plot we include fake images from the GAN (left) and fake images from the GAN augmented with FID (right), each row correspond to one repetition of the experiment. 
%We found no signs of generator cheating. 

\textbf{Experimental Conclusion. } 
The generators that use FID as an additional loss get better validation FID in all~experiments. 
This~raises an important question: Can optimization of FID as a loss ``improve'' generated images?
%Does the generators improve FID by producing ``better'' images? %better validation FID imply that the generative models improved? 

We address this question in \Cref{sec:finetune}. 
The remainder of this section, \Cref{subsec:advice}, presents further details regarding the experiments described in this~section. 

%\clearpage 
\subsection{Experimental Details}\label{subsec:advice}
%This section presents a few tricks we found to be vital in FID training. 

\paragraph{Training. }
GANs have two neural networks, a generator $G$ and a discriminator $D$. 
The generator is trained to make fake samples $F$ that minimize the discriminator loss $L_\disc(F)$, 
\begin{equation}
    \min L_\disc(F).
\end{equation}
The generator is often evaluated by computing FID between $10^4$ real images $X$ and $10^4$ fake images $F$, 
\begin{equation}
    \fid (X, F).
\end{equation}
We add FID to the generators loss, %In particular, we add FID between all training samples $X_{train}$ and a mini-batch of fake samples $f$. 
\begin{equation}
    \min L_\disc (F) + \fid (X, F).
\end{equation}
%We augment the loss function with FID between all validation samples $X_{\text{val}}$ and 10000 fake samples $F_{10^4}$. 
%\begin{equation}\label{equ:nllfid}
%    \min \mathbb{E}_{f\sim G}[\left[ L_\disc (f)] + \fid(X_{\text{val}}, F_{10^4})
%\end{equation} 
The generators then jointly minimize $L_\disc$ and FID.  
In all experiments, we optimized the joint loss function using gradient descent with a mini-batch of 128 fake examples. 
Notably, FastFID allows us to efficiently compute FID between the entire training data $X$ and a mini-batch of fake samples. 
In turn, we do not sample a mini-batch from the training data, which removes variance caused by sampling. 

The authors of FID suggest \emph{evaluating} FID with at least $10^4$ fake samples (FID$_{10^4}$). 
One might therefore be concerned that a mini-batch with $128$ fake samples (FID$_{128}$) is insufficient for \emph{training}.
In all three experiments, we saw training to minimize FID$_{128}$ consistently yielded smaller FID$_{10^4}$. 
This leads us to conclude a batch size of $128$ fake samples is sufficient when FID is used for \emph{training}. 

\paragraph{Scaling Loss.} 
The discriminator loss $L_\disc$ usually lies within $[-10,10]$, much smaller than FID$_{128}$, which typically lies within $[0, 500]$. 
This causes the gradients from FID$_{128}$ to be up to $50\times$ larger than the gradients from $L_\disc$, which subsequently breaks training. 
We circumvented this issue by adaptively scaling FID to match the discriminator loss~$L_\disc$. 
\begin{equation}\label{equ:backpropmemlimit}
    \text{scaled loss} =L_\disc + \fid \; / \underbrace{\fid \cdot L_\disc}_{\text{without gradients}}
\end{equation}
Training SAGAN was initially unstable, we found that further dividing the FID loss by 2 improved training stability. 

\paragraph{Computing FID. } 
FID is computed differently by PyTorch \cite{pytorch} and TensorFlow \cite{tf}. 
This is caused by architectural differences in the implementation of the Inception network, e.g., TensorFlow uses mean pooling while PyTorch use max pooling. 
These issues were fixed in an implementation by \cite{fidfix}, which we use to compute FID in the SNGAN and SAGAN experiment. 
To test whether the observed improvements were dependent to the specifics of the Inception architecture, we used the following PyTorch implementation in the DCGAN experiments: 

\href{https://github.com/hukkelas/pytorch-frechet-inception-distance}{github.com/hukkelas/pytorch-frechet-inception-distance}

\paragraph{Train and Validation Sets. } % TODO
\cite{fid} provide precomputed Inception statistics. 
% On some datasets the statistics are computed on training data, and, for other datasets, the statistics are computed on validation set.  
On some datasets the statistics are computed on the training data, while on others, they are computed on the validation data.  
Since we optimize FID on training data we report FID on validation data. 

\paragraph{GPU Memory.} \label{sec:exp_details}
The joint loss \Cref{equ:backpropmemlimit} requires us to backpropagate through both $L_\disc$ and FID, which increases peak memory consumption. 
We mitigate this issue with two separate backward passes for $L_\disc$ and~FID. 

The open-source implementation of SAGAN used two GPUs to reach batch size 128. 
Due to hardware limitations, we had to fit SAGAN on a single GPU.
However, SAGAN took up all 11 GB of our GPUs memory at batch size 64. 
To keep batch size 128 on a single GPU we used gradient checkpointing \cite{gradientcheckpointing} and 16-bit precision.  

\paragraph{Backpropagation and Eigenvalues.} \Cref{algo:pseudocode} needs to backpropagate through \textsc{torch.eig(M)}. % for $M=C_1^TC_2C_2^TC_1$.
At the time of writing, backpropagation through \textsc{torch.eig} is not supported in the stable release (PyTorch v1.7.1). 
Since $M$ is symmetric $M=M^T$ one can use \textsc{torch.symeig} which does support backpropagation. 
One can also use \textsc{torch.svd} to compute singular values, since $M$ is positive semi-definite the eigenvalues and singular values are equal $\lambda_i(M)=\sigma_i(M)$.
Alternatively, the unstable PyTorch 1.8 does support backpropagation through \textsc{torch.eig}.

\begin{figure}[h!]
    \centering
    \begin{tabular}{l l}
         (a) &   \includegraphics[width=0.35\textwidth]{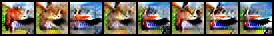}\\
         (b) & \includegraphics[width=0.35\textwidth]{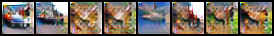}
    \end{tabular}
    \caption{(a) Glow trained to minimize FID produce samples with ``unrealistic'' artifacts. (b) Removing the discriminator from SNGAN also leads to ``unrealistic'' artifacts. }
    \label{fig:glow}
\end{figure}

\paragraph{Other Generative Models.} Normalizing Flows (NFs) \cite{nf} are generative models with many desirable properties, however, they sometimes attain poor~FID. 
The poor FID motivated us to train the NF called Glow \cite{glow} to minimize FID and negative log-likelihood.
The resulting Glow produced samples with ``unrealistic'' artifacts, see~\Cref{fig:glow}(a). 

The artifacts raise an interesting question: why does FID training cause Glow to produce artifacts but not SNGAN$_\fid$?
We~hypothesize that the discriminator learns to detect the ``unrealistic'' artifacts and penalizes the generator. %  appropriately. 
If this hypothesis is true, we would expect SNGAN$_\fid$ to produce ``unrealistic'' artifacts if we removed the discriminator. 
Indeed, if we train SNGAN$_\fid$ as in the previous section, but remove the discriminator after 1000 steps, the generator starts producing ``unrealistic'' artifacts, see \Cref{fig:glow}(b). 

\section{Can FID Loss Improve Generated Images?}
\label{sec:finetune} 
This section explores whether FID as a loss function can improve generated fake images. 
The experimental~setup: 
we~use~BigGAN pretrained\footnote{\href{https://github.com/ajbrock/BigGAN-PyTorch}{https://github.com/ajbrock/BigGAN-PyTorch}} for $10^5$ iterations and train the generator to minimize FID \emph{without its~discriminator}. 
The discriminator is discarded to ensure that changes in the fake images are due to the FID loss and \emph{not} the discriminator. 

\textbf{Goal.} 
A generator might improve the FID loss by changing a few pixels of the generated images to fool FID.\footnote{Fooling FID is similar to adversarial examples. }%FastFID can make adversarial examples for FID, see supplementary material. }
%  (or other imperceptible changes).
~Our~goal is to explore whether the FID loss leads the generator to ``fool FID''
or ``improve the fake images.''

\textbf{Observations. } We track how 64 samples change during training. 
All samples had perceptible changes ruling out ``few pixel changes'' (see supplementary material). 
We comment on two insightful samples below, which both demonstrate the addition of features like ears, eyes~or~heads. 

\begin{figure}[h]
    \centering
    \includegraphics[width=.49\textwidth]{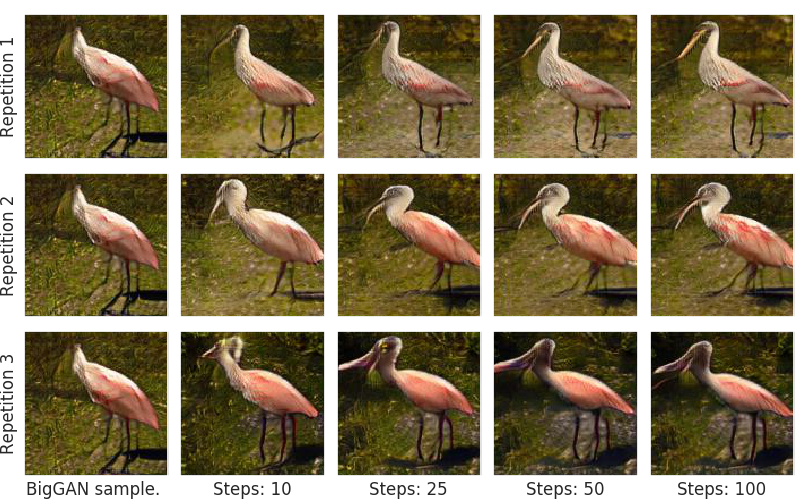}  
    \caption{Visualization of changes to samples while FID is minimized. Each row corresponds to a repetition of the experiment. Each column shows samples after a given number of steps. }
    \label{fig:bird}
\end{figure}

\Cref{fig:bird} contains BigGAN samples of a bird, each row corresponds to a repetition of the experiment. The left-most column shows the original bird generated by the pretrained BigGAN. The following columns demonstrate how the bird changes as FID is minimized. Notably, the initial bird lacks a beak. In all repetitions, we found that minimizing FID made the generator add a ``beak''-like feature. 

\begin{figure}[h]
    \centering
    \includegraphics[width=0.49\textwidth]{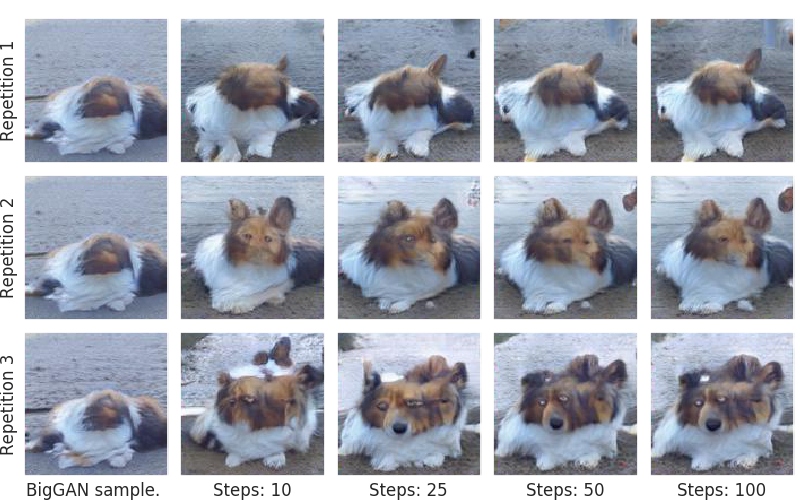}
    \caption{Visualization of changes to samples while FID is minimized. Each row corresponds to a repetition of the experiment. Each column shows samples after a given number of steps. }
    \label{fig:dog2}
\end{figure}

\Cref{fig:dog2} contains samples of a dog, where the initial dog has no head. % . The initial dog has no head. %in the initial BigGAN sample has no head. 
In all repetitions, we found that minimizing FID made the generator add features like ears, eyes or heads. 

\textbf{Conclusion. } FID minimization \emph{can} improve the fake images with the addition of features like ears, eyes or heads.
This demonstrates that FID minimization does not necessarily lead the generator to ``fool FID'' and \emph{can} (in some cases) ``improve the fake images.'' 


CODE: \verb^code.ipynb^ (one click to run in Google Colab).

\subsection{Further details}
\paragraph{Understanding FID. }
Some of the dogs in \Cref{fig:dog2} have two heads. 
Such behavior might be explained by carefully inspecting the FID equation. 
\begin{equation} 
      \underbrace
      {||\mu_1 - \mu_2||_2^2}_{\text{mean difference } \Delta_\mu}
      + 
      \underbrace
      {\tr(\Sigma_1) + \tr(\Sigma_2) - 2 \tr( \sqrt{\Sigma_1\Sigma_2})}_{\text{covariance difference } \Delta_\Sigma}.
\end{equation} 
The term $\Delta_\mu$ penalizes the differences between the average~Inception activations of the real data and the fake~data. 
Suppose~the entry $\mu_1^\text{head}$ and the entry $\mu_2^\text{head}$ contain the average number of heads detected by the Inception network.~\footnote{\cite{neurons} find high-level neurons activate based on complex features or whole objects like ``head`` or ``eye.''} 
Then $(\mu_1^\text{head}-\mu_2^\text{head})^2$ measures a difference in the average number of heads detected between the fake and real data. 
If the average number of heads in the fake data $\mu_1^\text{head}$ is smaller than the average number for the real data $\mu_2^\text{head}$, then $\Delta_\mu$ may encourage the generator to produce more heads. 
Notably, $\Delta_\mu$ might be indifferent to whether the heads are attached to a single dog. %to~one~dog. 

The term $\Delta_\Sigma$ penalizes the differences between the covariance of Inception activations of the real data and the fake~data. 
Suppose~the entry $\mu_i^\text{eye}$ contains the average number of eyes detected by the Inception network. 
Then,~$\Sigma_i^{\text{head, eye}}$ would measure the co-variance of heads and eyes. 

Altogether, it seems that FID incentivizes the generator to generate images where features (like heads and eyes) occur and co-occur in the same way as in training data. 
We~hypothesize this incentive is the reason why the FID loss can improve the generated samples. 

\paragraph{Experimental Notes.} 
The BigGAN model was pretrained for $10^5$ iterations with batch size 2048 using 8 GPUs (Tesla V100) for a total of 128 GB memory. 
We had access to a single Tesla P100 with 16 GB memory through Google Colab: 
\href{https://colab.research.google.com}{www.colab.research.google.com}.

To fit everything on a single 16 GB GPU, we used gradient checkpointing and half-precision training as in \Cref{sec:exp_details}. 
Furthermore, we accumulated gradients from batches of size 100 over 10 iterations to reach a total batch size of 1000. 
This still reduced the batch size from 2048 to 1000, which we corrected for by dividing the learning rate by two. 

To stabilize training, we fixed the latent vectors and only optimize the first two layers of the generator. 
Note that it is easier for the generator to ``fool FID'' for a fixed set of latent~vectors instead of $z\sim N(0, I)$ because it can take all values in~$\mathbb{R}^d$. 

%The fixed latent vectors should, in theory, make the experiment~deterministic. 
%However, we found repeating the experiment lead to different~results. 
%We suspect these differences are caused by the non-deterministic way GPUs evaluates, e.g., floating point addition. 

\section{Related Work}
The literature contains several methods for evaluating generative models. 
An overview can be found in \cite{overview} which reviews a total of 29 different methods.
This article concerns FID, which has arguably become the standard for benchmarking generative models in computer vision. %

\label{sec:rw}

\subsection{Fréchet Inception Distance}
\cite{fid} introduced FID and demonstrated that FID is consistent with increasing levels of disturbances and human judgment. 
To justify FID, the authors assume that the Inception activations are normally distributed. 
Whether this assumption holds or not, does not change the fact that the computed FID was consistent with increased levels of disturbances and human judgment.

The authors provide precomputed Inception statistics for five datasets\footnote{\label{fn:fid}\href{https://github.com/bioinf-jku/TTUR}{https://github.com/bioinf-jku/TTUR} \label{fn}}. 
Some use validation sets and others do not. 
This has, understandably, caused a bit of confusion, with some work reporting training FID and others reporting validation FID. 
We report validation FID since we explicitly optimize training FID.

The authors state\footnotemark[\value{footnote}] ``The number of samples should be greater than 2048. Otherwise $\Sigma$ is not full rank resulting in complex numbers and \verb!NaN!s by calculating the square root.'' 
We compute $\tr(\sqrt{\Sigma})$ directly without explicitly computing $\sqrt{\Sigma}$, which works even when $\Sigma$ has low rank. 
Our algorithm even becomes faster when the rank of $\Sigma$ decreases.

To the best of our knowledge, no previous work use FID as a loss function. 
We suspect the main obstacle preventing this in prior work has been the slow computation of FID, an issue FastFID mitigates. 

\subsection{Faster FID Computations}
To the best of our knowledge, there exist no published articles which reduce the asymptotic time complexity of FID. 
However, one implementation\footnote{\href{https://github.com/tensorflow/tensorflow/blob/00fad90125b18b80fe054de1055770cfb8fe4ba3/tensorflow/contrib/gan/python/eval/python/classifier_metrics_impl.py}
{
https://github.com/tensorflow/tensorflow/blob/00fad90125b18\\b80fe054de1055770cfb8fe4ba3/tensorflow/contrib/gan/python/eval\\/python/classifier\_metrics\_impl.py
}
} contains a few tricks. 
In this implementation, the function at \href{https://github.com/tensorflow/tensorflow/blob/00fad90125b18b80fe054de1055770cfb8fe4ba3/tensorflow/contrib/gan/python/eval/python/classifier_metrics_impl.py#L525}{line 525} reduces the time complexity from $O(d^3+d^2m)$ to $O(dm)$ by computing only $||\mu_1-\mu_2||_2^2$, omitting the trace term $\tr(\Sigma_1+\Sigma_2-2\sqrt{\Sigma_1\Sigma_2})$.   
Another function at \href{https://github.com/tensorflow/tensorflow/blob/00fad90125b18b80fe054de1055770cfb8fe4ba3/tensorflow/contrib/gan/python/eval/python/classifier_metrics_impl.py#L582}{line 582} incorporates the covariance matrices while retaining the $O(dm)$ time complexity, by only using the diagonal entries of the covariance matrices. 
Yet another function at \href{https://github.com/tensorflow/tensorflow/blob/00fad90125b18b80fe054de1055770cfb8fe4ba3/tensorflow/contrib/gan/python/eval/python/classifier_metrics_impl.py#L411}{line 411} computes $\tr(\sqrt{\Sigma_1\Sigma_2})$ by computing an eigendecomposition of two related matrices. %by an eigendecomposition. % computing the eigenvalues of $\sqrt{\Sigma_1}\Sigma_2\sqrt{\Sigma_1}$. 
%This has the advantage that $\Sigma_1$ is symmetric so one can compute $\sqrt{\Sigma_1}$ through its eigendecomposition.
%Such approach is not possible for $\Sigma_1\Sigma_2$ which, in general, is not symmetric. 
Their derivations have some similarities to our derivations, but they do not exploit low rank (small $m$) and take $O(d^3+d^2m)$ time instead of $O(d^2m + m^3)$. 
That said, eigendecompositions can be used to improve the baseline in \Cref{subsec:ouralgo} by a constant factor. 
Relative to such an improved baseline, in the example of \Cref{subsec:pracspeedup}, FastFID will remove roughly $13$ days of ``wasted'' computation instead of $130$ days.  %(see example in \Cref{subsec:pracspeedup}). 

\cite{approxsqrtm} present an algorithm that approximates a matrix square root, which is much faster than \sqrtm. 
Notably,~this approximation is used in the open-source implementations of BigGAN\footnotemark[10] to provide a fast FID approximation. 

The above algorithms all differ from FastFID, by either approximating FID, computing something entirely different or having an $O(d^3+d^2m)$ time complexity.%since they either approximate FID or compute something completely different.  % big-O 
%In theory, FastFID provably computes exactly the same result as \sqrtm. %In practice, we found that FastFID exhibits less numerical errors than \sqrtm.

%\subsection{Adversarial Examples} 
%The Inception Distance (ID) \cite{inceptiondistance} was widely used before the introduction of the FID. 
%Previous work has successfully constructed adversarial examples against ID \cite{adversarialinception}. 
%To the best of our knowledge, such adversarial examples have not been constructed for the FID. 
%FastFID allows constructing such adversarial examples efficiently by simple gradient descent as presented in \Cref{sec:adv}. 

\section{Conclusion}
\label{sec:conclusion}
Computing gradients through FID with automatic differentiation can increase training time by many days. 
%FastFID circumvents the increase in computation time by simplifying the computation through eigenvalue~considerations. 
FastFID computes the needed gradients efficiently by utilizing eigenvalue~considerations. 
This allows us to use FastFID to train GANs with FID as a loss function. 
Such training consistently improves the validation FID for different GANs on different datasets. 
We attempt to investigate whether FID as a loss encourages the generator to ``fool FID'' or generate better images. 
We find some evidence that suggests FID as a loss function \emph{can} improve the samples of a generator, however, understanding when this happens and when it does not remains unclear, and should be the focus of further research.

\paragraph{Remark. }FID can now be optimized directly. To allow fair comparisons in future research, we recommend future researchers explicitly denote whenever a generative model explicitly optimize FID, e.g., SNGAN$_\fid$ or SNGAN$^\fid$.  

\bibliography{bib}

\begin{thebibliography}{29}
\providecommand{\natexlab}[1]{#1}
\providecommand{\url}[1]{\texttt{#1}}
\expandafter\ifx\csname urlstyle\endcsname\relax
  \providecommand{\doi}[1]{doi: #1}\else
  \providecommand{\doi}{doi: \begingroup \urlstyle{rm}\Url}\fi

\bibitem[Abadi et~al.(2015)Abadi, Agarwal, Barham, Brevdo, Chen, Citro,
  Corrado, Davis, Dean, Devin, Ghemawat, Goodfellow, Harp, Irving, Isard, Jia,
  Jozefowicz, Kaiser, Kudlur, Levenberg, Man\'{e}, Monga, Moore, Murray, Olah,
  Schuster, Shlens, Steiner, Sutskever, Talwar, Tucker, Vanhoucke, Vasudevan,
  Vi\'{e}gas, Vinyals, Warden, Wattenberg, Wicke, Yu, and Zheng]{tf}
Abadi, M., Agarwal, A., Barham, P., Brevdo, E., Chen, Z., Citro, C., Corrado,
  G.~S., Davis, A., Dean, J., Devin, M., Ghemawat, S., Goodfellow, I., Harp,
  A., Irving, G., Isard, M., Jia, Y., Jozefowicz, R., Kaiser, L., Kudlur, M.,
  Levenberg, J., Man\'{e}, D., Monga, R., Moore, S., Murray, D., Olah, C.,
  Schuster, M., Shlens, J., Steiner, B., Sutskever, I., Talwar, K., Tucker, P.,
  Vanhoucke, V., Vasudevan, V., Vi\'{e}gas, F., Vinyals, O., Warden, P.,
  Wattenberg, M., Wicke, M., Yu, Y., and Zheng, X.
\newblock {TensorFlow}: {L}arge-{S}cale {M}achine {L}earning on {H}eterogeneous
  {S}ystems, 2015.
\newblock URL \url{http://tensorflow.org/}.
\newblock Software available from tensorflow.org.

\bibitem[Bj{\"o}rck \& Hammarling(1983)Bj{\"o}rck and Hammarling]{sqrtm}
Bj{\"o}rck, {\AA}. and Hammarling, S.
\newblock A {S}chur {M}ethod for the {S}quare {R}oot of a {M}atrix.
\newblock \emph{Linear algebra and its applications}, 1983.

\bibitem[Borji(2019)]{overview}
Borji, A.
\newblock Pros and {C}ons of {GAN} {E}valuation {M}easures.
\newblock \emph{Computer Vision and Image Understanding}, 179:\penalty0 41--65,
  2019.

\bibitem[Brock et~al.(2019)Brock, Donahue, and Simonyan]{biggan}
Brock, A., Donahue, J., and Simonyan, K.
\newblock Large {S}cale {GAN} {T}raining for {H}igh {F}idelity {N}atural
  {I}mage {S}ynthesis.
\newblock In \emph{International Conference on Learning Representations}, 2019.
\newblock URL \url{https://openreview.net/forum?id=B1xsqj09Fm}.

\bibitem[Chen et~al.(2016)Chen, Xu, Zhang, and Guestrin]{gradientcheckpointing}
Chen, T., Xu, B., Zhang, C., and Guestrin, C.
\newblock {Training Deep Nets with Sublinear Memory Cost}.
\newblock abs/1604.06174, 2016.

\bibitem[Chen et~al.(2015)Chen, Wilson, Tyree, Weinberger, and Chen]{compress}
Chen, W., Wilson, J., Tyree, S., Weinberger, K., and Chen, Y.
\newblock Compressing {N}eural {N}etworks with the {H}ashing {T}rick.
\newblock In \emph{International conference on machine learning}, pp.\
  2285--2294, 2015.

\bibitem[Deadman et~al.(2012)Deadman, Higham, and Ralha]{scipysqrtm}
Deadman, E., Higham, N.~J., and Ralha, R.
\newblock Blocked {S}chur {A}lgorithms for {C}omputing the {M}atrix {S}quare
  {R}oot.
\newblock In \emph{International Workshop on Applied Parallel Computing}, pp.\
  171--182. Springer, 2012.

\bibitem[Deng et~al.(2009)Deng, Dong, Socher, Li, Li, and Fei-Fei]{imagenet}
Deng, J., Dong, W., Socher, R., Li, L.-J., Li, K., and Fei-Fei, L.
\newblock {ImageNet: A Large-Scale Hierarchical Image Database}.
\newblock In \emph{CVPR09}, 2009.

\bibitem[Dinh et~al.(2015)Dinh, Krueger, and Bengio]{nf}
Dinh, L., Krueger, D., and Bengio, Y.
\newblock {NICE:} {N}on-{L}inear {I}ndependent {C}omponents {E}stimation.
\newblock In \emph{{ICLR} (Workshop)}, 2015.

\bibitem[Dowson \& Landau(1982)Dowson and Landau]{wasserstein}
Dowson, D. and Landau, B.
\newblock The {F}r{\'e}chet {D}istance between {M}ultivariate {N}ormal
  {D}istributions.
\newblock \emph{Journal of multivariate analysis}, 1982.

\bibitem[Goodfellow et~al.(2014)Goodfellow, Pouget-Abadie, Mirza, Xu,
  Warde-Farley, Ozair, Courville, and Bengio]{gans}
Goodfellow, I., Pouget-Abadie, J., Mirza, M., Xu, B., Warde-Farley, D., Ozair,
  S., Courville, A., and Bengio, Y.
\newblock Generative {A}dversarial {N}ets.
\newblock In \emph{NIPS}, 2014.

\bibitem[Heusel et~al.(2017)Heusel, Ramsauer, Unterthiner, Nessler, and
  Hochreiter]{fid}
Heusel, M., Ramsauer, H., Unterthiner, T., Nessler, B., and Hochreiter, S.
\newblock {GAN}s {T}rained by a {T}wo {T}ime-{S}cale {U}pdate {R}ule {C}onverge
  to a {L}ocal {N}ash {E}quilibrium.
\newblock In \emph{Advances in neural information processing systems}, pp.\
  6626--6637, 2017.

\bibitem[Karras et~al.(2020)Karras, Aittala, Hellsten, Laine, Lehtinen, and
  Aila]{nvidia}
Karras, T., Aittala, M., Hellsten, J., Laine, S., Lehtinen, J., and Aila, T.
\newblock {Training Generative Adversarial Networks with Limited Data}.
\newblock \emph{NeurIPS}, 2020.

\bibitem[Kingma \& Dhariwal(2018)Kingma and Dhariwal]{glow}
Kingma, D.~P. and Dhariwal, P.
\newblock Glow: Generative {F}low with {I}nvertible 1x1 {C}onvolutions.
\newblock In \emph{NeurIPS}, 2018.

\bibitem[Krizhevsky(2009)]{cifar}
Krizhevsky, A.
\newblock Learning {M}ultiple {L}ayers of {F}eatures {F}rom {T}iny {I}mages.
\newblock Technical report, 2009.

\bibitem[Lin \& Maji(2017)Lin and Maji]{approxsqrtm}
Lin, T.-Y. and Maji, S.
\newblock {Improved Bilinear Pooling with CNNs}.
\newblock In \emph{British Machine Vision Conference (BMVC)}, 2017.

\bibitem[Liu et~al.(2015)Liu, Luo, Wang, and Tang]{celeba}
Liu, Z., Luo, P., Wang, X., and Tang, X.
\newblock Deep {L}earning {F}ace {A}ttributes in the {W}ild.
\newblock In \emph{Proceedings of International Conference on Computer Vision
  (ICCV)}, December 2015.

\bibitem[Miyato et~al.(2018)Miyato, Kataoka, Koyama, and Yoshida]{sngan}
Miyato, T., Kataoka, T., Koyama, M., and Yoshida, Y.
\newblock Spectral {N}ormalization for {G}enerative {A}dversarial {N}etworks.
\newblock In \emph{ICLR}, 2018.

\bibitem[Mordvintsev et~al.(2015)Mordvintsev, Olah, and Tyka]{neurons}
Mordvintsev, A., Olah, C., and Tyka, M.
\newblock {Inceptionism: Going Deeper into Neural Networks}, 2015.

\bibitem[Nakatsukasa(2019)]{eigfact}
Nakatsukasa, Y.
\newblock The {L}ow-{R}ank {E}igenvalue {P}roblem, 2019.

\bibitem[Paszke et~al.(2019)Paszke, Gross, Massa, Lerer, Bradbury, Chanan,
  Killeen, Lin, Gimelshein, Antiga, Desmaison, Kopf, Yang, DeVito, Raison,
  Tejani, Chilamkurthy, Steiner, Fang, Bai, and Chintala]{pytorch}
Paszke, A., Gross, S., Massa, F., Lerer, A., Bradbury, J., Chanan, G., Killeen,
  T., Lin, Z., Gimelshein, N., Antiga, L., Desmaison, A., Kopf, A., Yang, E.,
  DeVito, Z., Raison, M., Tejani, A., Chilamkurthy, S., Steiner, B., Fang, L.,
  Bai, J., and Chintala, S.
\newblock Pytorch: An {I}mperative {S}tyle, {H}igh-{P}erformance {D}eep
  {L}earning {L}ibrary.
\newblock In \emph{NeurIPS}, 2019.

\bibitem[Polykovskiy et~al.(2018)Polykovskiy, Zhebrak, Sanchez-Lengeling,
  Golovanov, Tatanov, Belyaev, Kurbanov, Artamonov, Aladinskiy, Veselov,
  Kadurin, Nikolenko, Aspuru-Guzik, and Zhavoronkov]{moses}
Polykovskiy, D., Zhebrak, A., Sanchez-Lengeling, B., Golovanov, S., Tatanov,
  O., Belyaev, S., Kurbanov, R., Artamonov, A., Aladinskiy, V., Veselov, M.,
  Kadurin, A., Nikolenko, S., Aspuru-Guzik, A., and Zhavoronkov, A.
\newblock {M}olecular {S}ets ({MOSES}): {A} {B}enchmarking {P}latform for
  {M}olecular {G}eneration {M}odels.
\newblock \emph{arXiv preprint arXiv:1811.12823}, 2018.

\bibitem[Preuer et~al.(2018)Preuer, Renz, Unterthiner, Hochreiter, and
  Klambauer]{fcd}
Preuer, K., Renz, P., Unterthiner, T., Hochreiter, S., and Klambauer, G.
\newblock Fr{\'e}chet {C}hem{N}et {D}istance: a {M}etric for {G}enerative
  {M}odels for {M}olecules in {D}rug {D}iscovery.
\newblock \emph{Journal of chemical information and modeling}, 58\penalty0
  (9):\penalty0 1736--1741, 2018.

\bibitem[Radford et~al.(2016)Radford, Metz, and Chintala]{dcgan}
Radford, A., Metz, L., and Chintala, S.
\newblock Unsupervised {R}epresentation {L}earning with {D}eep {C}onvolutional
  {G}enerative {A}dversarial {N}etworks.
\newblock In \emph{4th International Conference on Learning Representations,
  {ICLR}}, 2016.

\bibitem[Seitzer(2020)]{fidfix}
Seitzer, M.
\newblock {PyTorch-FID: FID Score for PyTorch}.
\newblock \url{https://github.com/mseitzer/pytorch-fid}, August 2020.
\newblock Version 0.1.1.

\bibitem[Szegedy et~al.(2015)Szegedy, Liu, Jia, Sermanet, Reed, Anguelov,
  Erhan, Vanhoucke, and Rabinovich]{inception}
Szegedy, C., Liu, W., Jia, Y., Sermanet, P., Reed, S., Anguelov, D., Erhan, D.,
  Vanhoucke, V., and Rabinovich, A.
\newblock Going {D}eeper with {C}onvolutions.
\newblock In \emph{Proceedings of the IEEE conference on computer vision and
  pattern recognition}, 2015.

\bibitem[Tran et~al.(2019)Tran, Tran, Nguyen, Yang, and Cheung]{saganfid}
Tran, N.-T., Tran, V.-H., Nguyen, N.-B., Yang, L., and Cheung, N.-M.
\newblock {Self-Supervised GAN: Analysis and Improvement With Multi-Class
  MiniMax Game}.
\newblock \emph{NeurIPS}, 2019.

\bibitem[{Virtanen} et~al.(2020){Virtanen}, {Gommers}, {Oliphant}, {Haberland},
  {Reddy}, {Cournapeau}, {Burovski}, {Peterson}, {Weckesser}, {Bright}, {van
  der Walt}, {Brett}, {Wilson}, {Jarrod Millman}, {Mayorov}, {Nelson}, {Jones},
  {Kern}, {Larson}, {Carey}, {Polat}, {Feng}, {Moore}, {Vand erPlas},
  {Laxalde}, {Perktold}, {Cimrman}, {Henriksen}, {Quintero}, {Harris},
  {Archibald}, {Ribeiro}, {Pedregosa}, {van Mulbregt}, and
  {Contributors}]{scipy}
{Virtanen}, P., {Gommers}, R., {Oliphant}, T.~E., {Haberland}, M., {Reddy}, T.,
  {Cournapeau}, D., {Burovski}, E., {Peterson}, P., {Weckesser}, W., {Bright},
  J., {van der Walt}, S.~J., {Brett}, M., {Wilson}, J., {Jarrod Millman}, K.,
  {Mayorov}, N., {Nelson}, A. R.~J., {Jones}, E., {Kern}, R., {Larson}, E.,
  {Carey}, C., {Polat}, {\.I}., {Feng}, Y., {Moore}, E.~W., {Vand erPlas}, J.,
  {Laxalde}, D., {Perktold}, J., {Cimrman}, R., {Henriksen}, I., {Quintero},
  E.~A., {Harris}, C.~R., {Archibald}, A.~M., {Ribeiro}, A.~H., {Pedregosa},
  F., {van Mulbregt}, P., and {Contributors}, S. .~.
\newblock {SciPy 1.0: Fundamental Algorithms for Scientific Computing in
  Python}.
\newblock \emph{Nature Methods}, 2020.

\bibitem[Zhang et~al.(2019)Zhang, Goodfellow, Metaxas, and Odena]{sagan}
Zhang, H., Goodfellow, I., Metaxas, D., and Odena, A.
\newblock {Self-Attention Generative Adversarial Networks}.
\newblock In \emph{ICML}, 2019.

\end{thebibliography}
\bibliographystyle{icml2021}

\clearpage 

\begin{figure*}[h!]
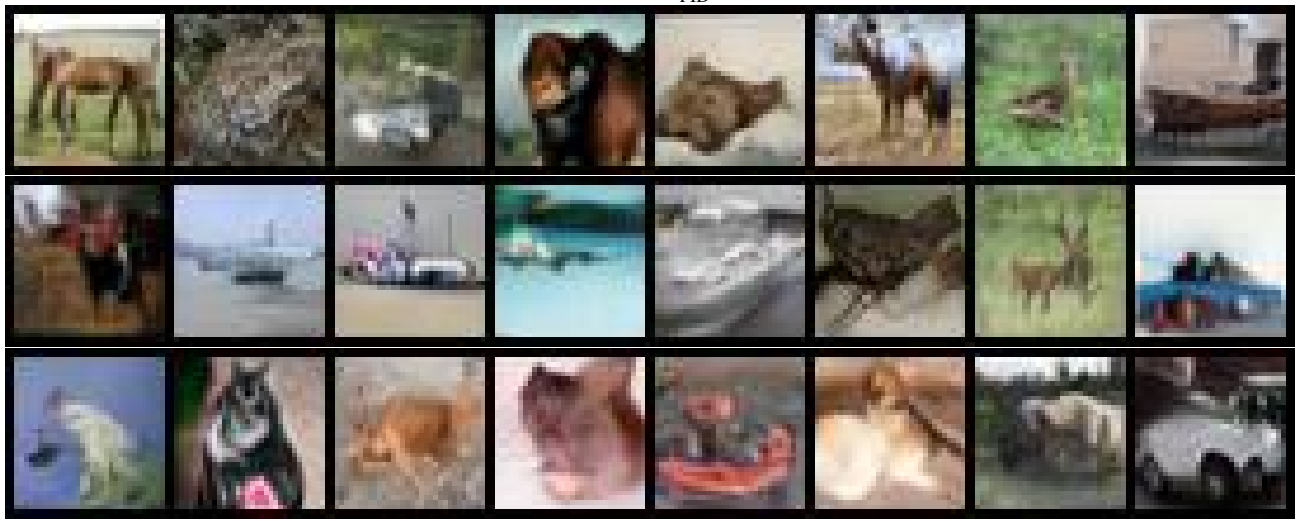

    \centering
    SNGAN$_\gan$\\
    \adjincludegraphics[trim={{0.0\width} 0 0 0}, clip, width=1\textwidth]{figures/cifar_samples/gan1.jpg}
    \adjincludegraphics[trim={{0.0\width} 0 0 0}, clip,width=1\textwidth]{figures/cifar_samples/gan2.jpg}
    \adjincludegraphics[trim={{0.0\width} 0 0 0}, clip,width=1\textwidth]{figures/cifar_samples/gan3.jpg}
    \\
    SNGAN$_\fid$\\
    \adjincludegraphics[trim={{0.0\width} 0 0 0}, clip,width=1\textwidth]{figures/cifar_samples/fid1.jpg}
    \adjincludegraphics[trim={{0.0\width} 0 0 0}, clip,width=1\textwidth]{figures/cifar_samples/fid2.jpg}
    \adjincludegraphics[trim={{0.0\width} 0 0 0}, clip,width=1\textwidth]{figures/cifar_samples/fid3.jpg}
    \caption{Bigger version of samples in \Cref{fig:sngan_train}.}
\end{figure*}
\clearpage 
\begin{figure*}[h]
    \centering
    DCGAN$_\gan$\\
    \adjincludegraphics[trim={{0.0\width} 0 0 0}, clip,width=1\textwidth]{figures/gan1.jpg}
    \adjincludegraphics[trim={{0.0\width} 0 0 0}, clip,width=1\textwidth]{figures/gan2.jpg}
    \adjincludegraphics[trim={{0.0\width} 0 0 0}, clip,width=1\textwidth]{figures/gan3.jpg}
    \\DCGAN$_\fid$\\
    \adjincludegraphics[trim={{0.0\width} 0 0 0}, clip,width=1\textwidth]{figures/fid1.jpg}
    \adjincludegraphics[trim={{0.0\width} 0 0 0}, clip,width=1\textwidth]{figures/fid2.jpg}
    \adjincludegraphics[trim={{0.0\width} 0 0 0}, clip,width=1\textwidth]{figures/fid3.jpg}
   \caption{Bigger version of samples in \Cref{fig:dcgan_train}. }
\end{figure*}
   
\clearpage 
\begin{figure*}[h]
    \centering
    SAGAN$_\gan$\\
    \adjincludegraphics[trim={{0.0\width} 0 0 0}, clip, width=1\textwidth]{figures/imagenet/gan1.jpg}
    \adjincludegraphics[trim={{0.0\width} 0 0 2}, clip, width=1\textwidth]{figures/imagenet/gan2.jpg}
    \adjincludegraphics[trim={{0.0\width} 0 0 0}, clip, width=1\textwidth]{figures/imagenet/gan3.jpg}
    \\SAGAN$_\fid$\\ 
    \adjincludegraphics[trim={{0.0\width} 0 0 0}, clip, width=1\textwidth]{figures/imagenet/fid1.jpg}
    \adjincludegraphics[trim={{0.0\width} 0 0 0}, clip, width=1\textwidth]{figures/imagenet/fid2.jpg}
    \adjincludegraphics[trim={{0.0\width} 0 0 0}, clip, width=1\textwidth]{figures/imagenet/fid3.jpg}
    \caption{Bigger version of samples in \Cref{fig:sagan_train}. }
    %\label{fig:sagan_train}
\end{figure*}

\end{document}